\documentclass[letterpaper, 10 pt, journal, twoside]{IEEEtran}
\usepackage{amsmath,amsfonts}
\usepackage{algorithmic}
\usepackage{algorithm}
\usepackage{array}
\usepackage{hyperref}
\usepackage[caption=false,font=small,labelfont=sf,textfont=sf]{subfig}
\usepackage{textcomp}

\usepackage{epstopdf}
\usepackage{stfloats}
\usepackage{url}
\usepackage{verbatim}
\usepackage{graphicx}
\usepackage{cite}
\usepackage{multirow}
\usepackage{booktabs}
\usepackage{ifthen}  
\usepackage{tabularx}
\usepackage{cleveref}
\usepackage{color}
\usepackage{transparent}
\graphicspath{{Diagrams/}}
\newcommand{\bftab}{\fontseries{b}\selectfont}
\usepackage[switch]{lineno}
\usepackage{flushend}
\usepackage{balance}
\usepackage{etoolbox}
\makeatletter
\patchcmd{\@makecaption}
  {\scshape}
  {}
  {}
  {}
\makeatletter
\patchcmd{\@makecaption}
  {\\}
  {.\ }
  {}
  {}
\makeatother

\usepackage{pgf}
\usepackage{pgfplots}  
\pgfplotsset{compat=1.18} 

\begin{document}
\bstctlcite{IEEEexample:BSTcontrol}
\title{\bf ForestVO: Enhancing Visual Odometry in Forest Environments through ForestGlue}

\author{
{Thomas Pritchard$^{*,1}$, Saifullah Ijaz$^{*,1}$,Ronald Clark$^{2}$, Basaran Bahadir Kocer$^{1,3}$}

\thanks{ 
\par $^{*}$ contributed equally
\par $^{1}$Computing Department, Imperial College London.
\par $^{2}$Department of Computer Science, University of Oxford.
\par $^{3}$School of Civil, Aerospace and Design Engineering, University of Bristol.
\par The proposed approach is available at: 
\par \url{https://github.com/AerialRoboticsGroup/forest-vo}}
}




\maketitle

\begin{abstract}
Recent advancements in visual odometry systems have improved autonomous navigation, yet challenges persist in complex environments like forests, where dense foliage, variable lighting, and repetitive textures compromise the accuracy of feature correspondences. To address these challenges, we introduce ForestGlue. ForestGlue enhances the SuperPoint feature detector through four configurations -- grayscale, RGB, RGB-D, and stereo-vision inputs -- optimised for various sensing modalities. For feature matching, we employ LightGlue or SuperGlue, both of which have been retrained using synthetic forest data. ForestGlue achieves comparable pose estimation accuracy to baseline LightGlue and SuperGlue models, yet require only 512 keypoints, just 25\% of the 2048 keypoints used by baseline models, to achieve an LO-RANSAC AUC score of 0.745 at a 10\textdegree\ threshold. With a 1/4 of the keypoints required, ForestGlue has the potential to reduce computational overhead whilst being effective in dynamic forest environments, making it a promising candidate for real-time deployment on resource-constrained platforms such as drones or mobile robotic platforms. By combining ForestGlue with a novel transformer based pose estimation model, we propose ForestVO, which estimates relative camera poses using the 2D pixel coordinates of matched features between frames. On challenging TartanAir forest sequences, ForestVO achieves an average relative pose error (RPE) of 1.09 m and kitti\_score of 2.33\%, outperforming direct-based methods such as DSO in dynamic scenes by 40\%, while maintaining competitive performance with TartanVO despite being a significantly lighter model trained on only 10\% of the dataset. This work establishes an end-to-end deep learning pipeline tailored for visual odometry in forested environments, leveraging forest-specific training data to optimise feature correspondence and pose estimation for improved accuracy and robustness in autonomous navigation systems.

\end{abstract}


\section{Introduction}
Forest environments compose a substantial part of the natural terrain throughout the world. In these settings, autonomous robots hold great potential for reducing direct human involvement with diverse applications in forestry-related tasks \cite{buchelt_exploring_2024,kocer2021forest,hauf2023learning,lan2024aerial}. Navigation systems commonly rely on GPS for localisation and path planning in structured environments \cite{patrik2019gnss}. However, dense canopy cover in forests renders GPS ineffective, posing significant challenges for autonomous forest navigation \cite{tian_search_2020}.

LiDAR-based odometry has been explored as an alternative to GPS for localisation in challenging environments \cite{lee2024lidar} by emitting laser pulses to measure distances and build high-resolution 3D maps. LIDAR is less affected by lighting variations, making it suitable for low-visibility conditions like forests. However, real-time processing of dense point clouds is computationally demanding. Additionally, LiDAR is sensitive to environmental interference from fog, rain, and airborne particles, causing beam scattering that reduces accuracy \cite{Dassot2011}. These limitations, coupled with the high cost and weight of LiDAR systems, make them less practical for lightweight, economical drones commonly used in forest exploration \cite{xiao2021optic,ho2022vision,kocer2022immersive,quan2023tree}.

\begin{figure}[t!]
    \centering
    \def\svgwidth{0.7\linewidth}
    \includegraphics[width=\linewidth]{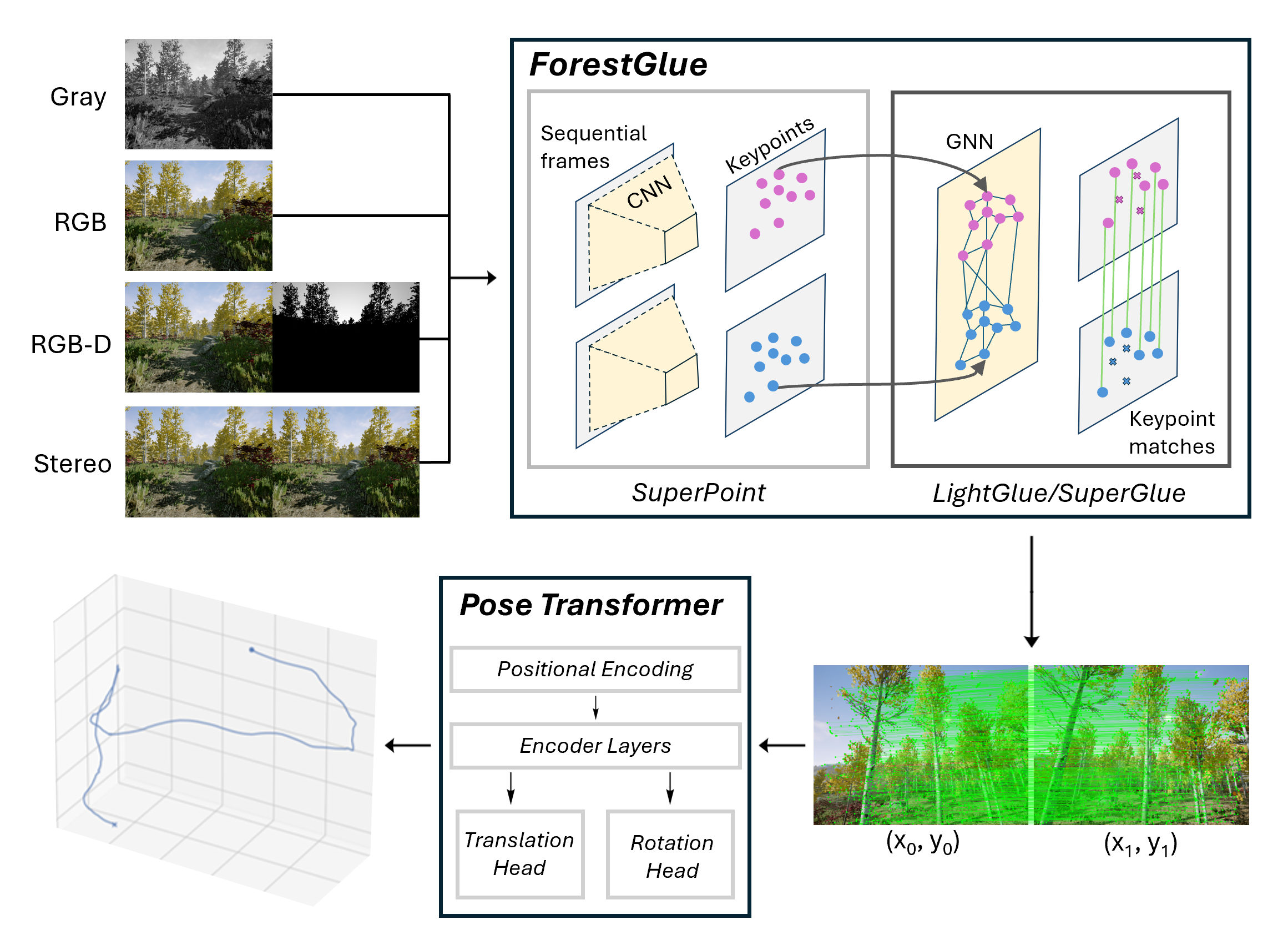}
    \caption{\textbf{ForestVO System Architecture:} ForestVO consists of ForestGlue and a deep-learning pose estimation model. Image pairs are passed through ForestGlue which performs feature detection using multi-modal SuperPoint and feature matching using either SuperGlue or LightGlue, which have been refined on forest-specific datasets. The matched 2D keypoint coordinates between sequential frames are passed into the pose transformer model, which estimates the relative camera pose by predicting the rotation matrix \( \mathbf{R} \) and translation vector \( \mathbf{T} \).    
    The relative poses are concatenated to generate a predicted trajectory for the sequence of input images.}
    \label{fig:sysarch}
    \vspace{-1em}
\end{figure}

In contrast, visual odometry (VO) provides a lightweight and cost-effective alternative by estimating a robot’s 3D motion using visual data captured by onboard cameras \cite{1315094}. Unlike LiDAR, VO systems avoid expensive sensors and are better suited for deployment on small, resource-constrained platforms such as drones or mobile robotic platforms \cite{aqel_review_2016}.
State-of-the-art VO methods achieve accurate localisation in structured environments, as demonstrated on benchmark vSLAM datasets such as KITTI, EUROC, and TUM RGB-D \cite{geiger_are_2012, burri_euroc_2016, sturm_benchmark_2012}. However, traditional geometric-based pose estimation methods used in VO are highly dependent on hand-crafted feature correspondences, which degrade in dynamic, unstructured settings like forests \cite{laina2024scalable}. To address these limitations, this work focuses on improving VO robustness in forest environments through domain-adapted feature-matching techniques that enhance keypoint detection and correspondence in challenging conditions.

Deep learning feature extractors and matchers have become increasingly prevalent to overcome the limitations of traditional techniques, achieving SOTA performance on image-matching tasks \cite{huang2024survey}. The ability to adapt to specific environments has demonstrated enhanced performance and robustness in image-matching tasks. Furthermore, deep learning in VO systems \cite{DBLP:journals/corr/abs-1709-08429} has been explored as an alternative end-to-end approach, directly regressing the relative camera poses from the input images, instead of geometric-based pose estimation methods. However, there is limited research on domain-specific VO for forest environments.

Therefore, we propose ForestVO, a deep learning-based VO framework for forest environments, utilising ForestGlue for feature correspondence. ForestGlue is a set of domain-adapted models for feature extraction and matching in forest environments. For feature extraction, we adapted the deep-learning SuperPoint network to be configured with one of four input modalities, grayscale, RGB, RGB-D, or stereo-vision \cite{detone2018superpoint}. Next, these keypoints are matched via SuperGlue \cite{sarlin2020superglue} or LightGlue \cite{lindenberger2306lightglue}, each retrained on synthetic TartanAir forest scenes to enhance robustness in dense, natural environments \cite{wang2020tartanair}.
The relative camera pose between frames is estimated using our ForestVO transformer model to directly regress the relative camera poses using the 2D-pixel coordinates of the matched keypoints across a pair of frames. The pose estimation model was also trained on scenes from TartanAir, including \textit{gascola}, \textit{seasons\_forest} and \textit{seasons\_forest\_winter} \cite{wang2020tartanair}.

\section{Related Work}
Identifying feature correspondences between image pairs is crucial for estimating 3D structures and camera poses in computer vision applications such as Simultaneous Localisation and Mapping (SLAM) \cite{durrant2006simultaneous} and Structure-from-Motion (SfM) \cite{sattler2016efficient}. Typically, this involves extracting sparse interest points from image pairs and matching local features based on high-dimensional representations of each feature's appearance \cite{tan2015feature}. However, in unstructured environments with variable illumination, viewpoint changes, blur, and occlusion, accurately matching features becomes challenging, as representations may be incomplete or indistinguishable between frames \cite{garforth2019visual,chen2019multi, chen2025imageimu}. This necessitates a feature matcher that can effectively reject outliers whilst reliably matching shared keypoints.

Traditional methods, such as SIFT \cite{lowe2004SIFT} and ORB \cite{ethan2011orb} rely on handcrafted descriptors and heuristic approaches to extract local features, which are then matched based on Euclidean distance or other similarity metrics \cite{tan2015feature}. While robust in structured environments, these approaches often struggle in complex, unstructured settings due to their reliance on fixed feature representations \cite{garforth2019visual,chen2019multi}. Nearest Neighbour (NN) is a typical traditional approach for keypoint matching, using the descriptors' Euclidean distance to determine correspondences \cite{huang2024survey, cover1967nearest}. To minimise false matches, a ratio test is applied, comparing the distance to the nearest neighbour (\(d_{NN}\)) with the distance to the second nearest neighbour (\(d_{NN2}\)). Matches are retained if the ratio between the two distances satisfies the condition:
\begin{align}
\frac{d_{NN2}}{d_{NN}} &< \text{threshold}
\end{align}

In contrast, deep learning-based methods for feature matching employ learned feature descriptors that adaptively model the visual context and structure of the scene \cite{huang2024survey}. Unlike traditional methods that rely on fixed descriptors, these approaches learn feature representations that are more flexible and can adjust to complex variations in visual data, such as changes in illumination, viewpoint, and occlusion. Feature correspondences are determined by optimising a learned distance metric:
\begin{align}
d_{\text{learned}}(f_i, f_j) &= \| f_i - f_j \|_2
\end{align}
where \( f_i \) and \( f_j \) are learned feature descriptors for keypoints in two images. This ability to learn context-specific features facilitates improved correspondence accuracy, particularly in scenarios with substantial scene structure variability \cite{sun2021loftr, rocco2020ncnet, huang2024survey}. 

SuperGlue \cite{sarlin2020superglue}, a deep neural network with a transformer backbone, leverages information from both images to reject outliers whilst maintaining discriminative accuracy. Notably, it has been successfully integrated into real-time SLAM systems, demonstrating its capability for real-time performance \cite{zhu2023novel}. Similarly, LightGlue \cite{lindenberger2306lightglue}, an adaptive variant of SuperGlue, enhances feature matching by dynamically adjusting to visual overlap and discriminative cues, making it particularly well-suited for latency-sensitive applications, and has also been implemented in real-time SLAM systems \cite{xu2024airslam,xiao2024sl}. However, both models were initially trained on the structured urban datasets Oxford-Paris 1M and MegaDepth \cite{radenovic2018revisitingoxford, li2018megadepth}. These models have yet to undergo extensive evaluation in unstructured environments, like forests, which present unique challenges and demand high reliability for critical applications in search and rescue or disaster management \cite{kanzaki2020uav,miyano2019utility}.

To bridge this gap, we have retrained SuperGlue and LightGlue on domain-specific data. Furthermore, we explored incorporating enhancements such as multi-modal inputs into SuperPoint and integrating an epipolar constraint in the loss function to improve feature-correspondence accuracy.

Classical VO/SLAM approaches rely on either direct or indirect methods. Indirect methods usually rely on hand-crafted feature methods, resulting in systems such as ORB-SLAM \cite{DBLP:journals/corr/Mur-ArtalMT15} that perform pose estimation by matching ORB features across frames before applying the PnP algorithm \cite{Lepetit2008}. On the other hand, direct methods such as DSO \cite{Engel-et-al-pami2018} rely on dense pixel matches via optical flow methods, aiming to estimate camera motion using the movements of the pixels between frames. Learning-based VO methods have also been explored, such as TartanVO \cite{DBLP:journals/corr/abs-2011-00359}, although they typically involve passing the images directly into the models to regress camera poses. Since learning-based feature methods such as SuperPoint and LightGlue have recently achieved SOTA performance on image-matching tasks, we have decided to follow the traditional pipeline of feature extraction, feature matching and pose estimation. However, we use separate deep learning models for each task in our ForestVO framework instead of the traditional geometric approaches or the image-based deep learning approaches.

\section{ForestGlue}
\subsection{Model Architectures}
We employed the original architectures of SuperGlue \cite{sarlin2020superglue} and LightGlue \cite{lindenberger2306lightglue} for retraining. We modified SuperPoint for RGB, RGB-D, and stereo input modalities by duplicating the pre-trained weights across the input layer \cite{detone2018superpoint}.
SuperGlue leverages a graph neural network with self- and cross-attention layers to iteratively refine keypoint descriptors and uses the Sinkhorn algorithm for optimal transport to identify correspondences. LightGlue, a lightweight alternative, employs the same backbone, but has adaptive mechanisms, including early termination and point pruning.

\subsection{Datasets}
The models were trained and evaluated on TartanAir and FinnForest. TartanAir is a synthetic dataset generated using Microsoft AirSim, providing a photo-realistic environment for SLAM and VO tasks \cite{wang2020tartanair}. The \emph{seasons\_forest} and \emph{seasons\_forest\_winter} subsets including 50K images were split into 40K/5K/5K for training, validation, and test sets. FinnForest is a real-world dataset collected in Finland boreal forests, offering high-resolution images with accurate ground truth pose data obtained through RTK GPS sensors \cite{ali2020finnforest}. 10K images from \emph{S01\_13Hz}, \emph{S02\_13Hz}, and \emph{W07\_10Hz} scenes represented summer and winter forests, providing a challenging and realistic test set.

\subsection{Homography and Depth Training}
The models trained from scratch were initially pre-trained on synthetic homographies, following the procedures outlined by Lindenberger et al. \cite{lindenberger2306lightglue}. To simulate sequential frames, homography transformations and photometric augmentations were applied. The transformation difficulty was set to 0.5, which linearly scales the magnitude of geometric distortions, reducing rotations, translations, and scaling to half of their maximum allowable range, with a maximum angle of 10°. This value was selected to balance variability and realism, to generate transformations that reflect plausible viewpoint changes encountered during UAV flights in forested areas. 
Ground truth correspondences were established by reprojecting keypoints using the homography transformation matrix, where keypoints falling within a 3-pixel radius were deemed positive matches.

We performed depth-based fine-tuning with TartanAir’s precise depth and pose data, projecting keypoints from one frame onto another using the relative pose; matches within a 3-pixel radius were classified as positive. Positive matches were filtered by colour similarity, excluding those with Euclidean colour difference over 10 to mitigate incorrect matches due to dynamic elements.

\subsection{Model Training Configuration}
We followed the training procedures outlined by the LightGlue and SuperGlue authors' \cite{sarlin2020superglue,lindenberger2306lightglue}. The homography models were initially trained from random weights for 40 epochs on 40K images from the TartanAir dataset, with a batch size of 64 and an initial learning rate of $1e^{-4}$, decaying by 0.8 after 20 epochs. For consequent depth-based training using sequential frames, the weights were either loaded from the authors' baseline models or homography training. Training and inference were performed on an HP Z640 with an Intel i7-7700K CPU and an NVIDIA GTX 1080 GPU (8GB VRAM). Due to computational constraints, we reduced the batch size to 8. To mitigate the noise from a smaller batch size, we lowered the initial learning rate to $5e^{-6}$ and applied gradient clipping. An exponential decay of 0.95 every epoch after 10 epochs was applied, with a total of 50 epochs. Unlike Lindenberger et al. \cite{lindenberger2306lightglue}, the top 512 keypoints were extracted instead of 2048 to encourage the model to effectively match with fewer keypoints. To further optimise training efficiency and manage memory usage, we utilised FlashAttention and gradient checkpointing.

We propose a novel loss function incorporating an epipolar error term weighted by match confidence to encourage geometric consistency in correspondence predictions. This epipolar hinge loss quantifies the deviation of predicted correspondences from the underlying epipolar geometry, providing a measure of geometric consistency. To mitigate the impact of confidently incorrect matches, the error is weighted by each match’s confidence score, thereby penalising high-confidence outliers. Minimising this weighted epipolar error during training encourages the model to produce geometrically accurate correspondences with more reliable confidence scores, thereby improving pose estimation, which is computed using the highest-confidence matches.

\begin{align}
    \tiny
\sum_{i=1}^{N} \min\left( \frac{1}{M_i} \sum_{j=1}^{M_i} \left( E_{ij} \cdot \frac{\log(1 + C_{ij})}{\max\left(\log(1 + C_{ij})\right)} \right), \alpha \cdot \mathcal{L}_{\text{LightGlue}} \right)
\end{align}

\textbf{Epipolar Hinge Loss:} The final loss is computed by adding a hinge term to the original LightGlue loss. \(\mathcal{L}_{\text{LightGlue}}\): LightGlue loss, \(N\): batch size, \(M_i\): valid match count for image \(i\), \(E_{ij}\): epipolar error for match \(j\) in image \(i\), and \(C_{ij}\): match confidence score. The normalised epipolar error was clipped to \(\alpha\), 0.2, of the original loss to mitigate exploding gradients.

\section{Pose Estimation Model}
\subsection{Model Architecture}
For our pose estimation model we leverage the transformer architecture, using the multi-head attention mechanism to attend to different parts of the input sequence, linking the spatial relationship between the coordinates of the matched keypoints to the movement of the camera position between frames. The model consists of a linear layer which projects the inputs to the dimensionality of the transformer encoder layers before applying positional encoding using sinusoidal functions to embed the positions of each keypoint in the sequence. Four transformer encoder layers are used, each containing four attention heads to understand how each keypoint has moved from one frame to the next. The output is then aggregated, taking the mean prediction based on the movements of each keypoint in the sample. The pose estimation transformer predictions are then separated into the 3D translation and 6D rotation components.

\subsection{Dataset}
We trained the pose estimation model using \textit{gascola}, \textit{seasons\_forest} and \textit{seasons\_forest\_winter} sequences in the TartanAir dataset, totalling 75000 images with an 80/10/10 split for training, validation and testing.
Image pairs are passed through ForestGlue (Grayscale SuperPoint + LightGlue)
to obtain the 2D coordinates of the matched keypoints in both frames, \((x0, y0)\) and \((x1, y1)\) which have the shape \((N, 2)\). The corresponding ground truth relative pose is obtained by computing the relative transformation between the absolute ground truth poses of the two frames. Each sample consists of the concatenated keypoints, of shape \((N, 4)\) as well as the ground truth relative translation and rotation. The translation component \([x,y,z]\) denotes the 3D relative translation between frames. The rotation component is converted to a 6D representation (the first two columns of a \(3\times3\) rotation matrix), a continuous format that is more suitable for neural networks. At inference, the original \(3\times3\) rotation matrix is recovered using a Gram-Schmidt like process \cite{DBLP:journals/corr/abs-1812-07035}.

\subsection{Training Setup}
We trained the pose estimation model with a batch size of 128, applying padding and masks to ensure uniform sequence lengths given varying keypoint counts between sequential frames.
We set the input dimensions to four to match the shape of the keypoint coordinates, the model's dimensionality to 128, and the feedforward dimension to 256, using a dropout of 0.1 for regularisation. We use the mean squared error (MSE) loss function with separate translation and rotation losses.
\begin{align}
\mathcal{L}_{\text{translation}} &= \text{MSE}(\hat{t}, t_{\text{rel}}) \\
\mathcal{L}_{\text{rotation}} &= \text{MSE}(\hat{r}, r_{\text{rel}}^{6\rm{D}}) \\
\mathcal{L}_{\text{total}} &= \mathcal{L}_{\text{translation}} + \beta \cdot \mathcal{L}_{\text{rotation}}
\end{align}
\textbf{Pose Estimation Loss:} The overall loss is computed by summing the relative translation and rotation losses. We balance the overall loss using a scale factor \(\beta\) = 100, since we found that the relative rotations between frames were approximately 100 times smaller than the translations in the TartanAir dataset. This prevents the model weights from updating mainly based on relative translation losses and ensures the model learns to correctly predict both the translation and rotation components of the relative camera poses.

We train the model for 28K steps, using the PyTorch Adam optimiser with a learning rate of $1e^{-3}$ and a weight decay of $1e^{-5}$. The model is very lightweight, containing only 531,721 trainable parameters. During inference the entire ForestVO framework (SuperPoint, LightGlue and the pose estimation model) requires only 1.2GB of VRAM. The hardware setup we used for inference and training includes a laptop containing an RTX 3080 GPU with 8GB VRAM, AMD Ryzen 7 6800HS CPU and 48GB RAM. For reference, ForestVO can process the \textit{seasonsforest\_example\_P002} sequence in the TartanAir dataset, which contains 301 images, in approximately 36.756 s.

\section{Experiments}
\subsection{Benchmarking Existing Models}
\begin{figure}[t!]
\vspace{-0.5em}
\centering
    \includegraphics[width=0.8\linewidth]{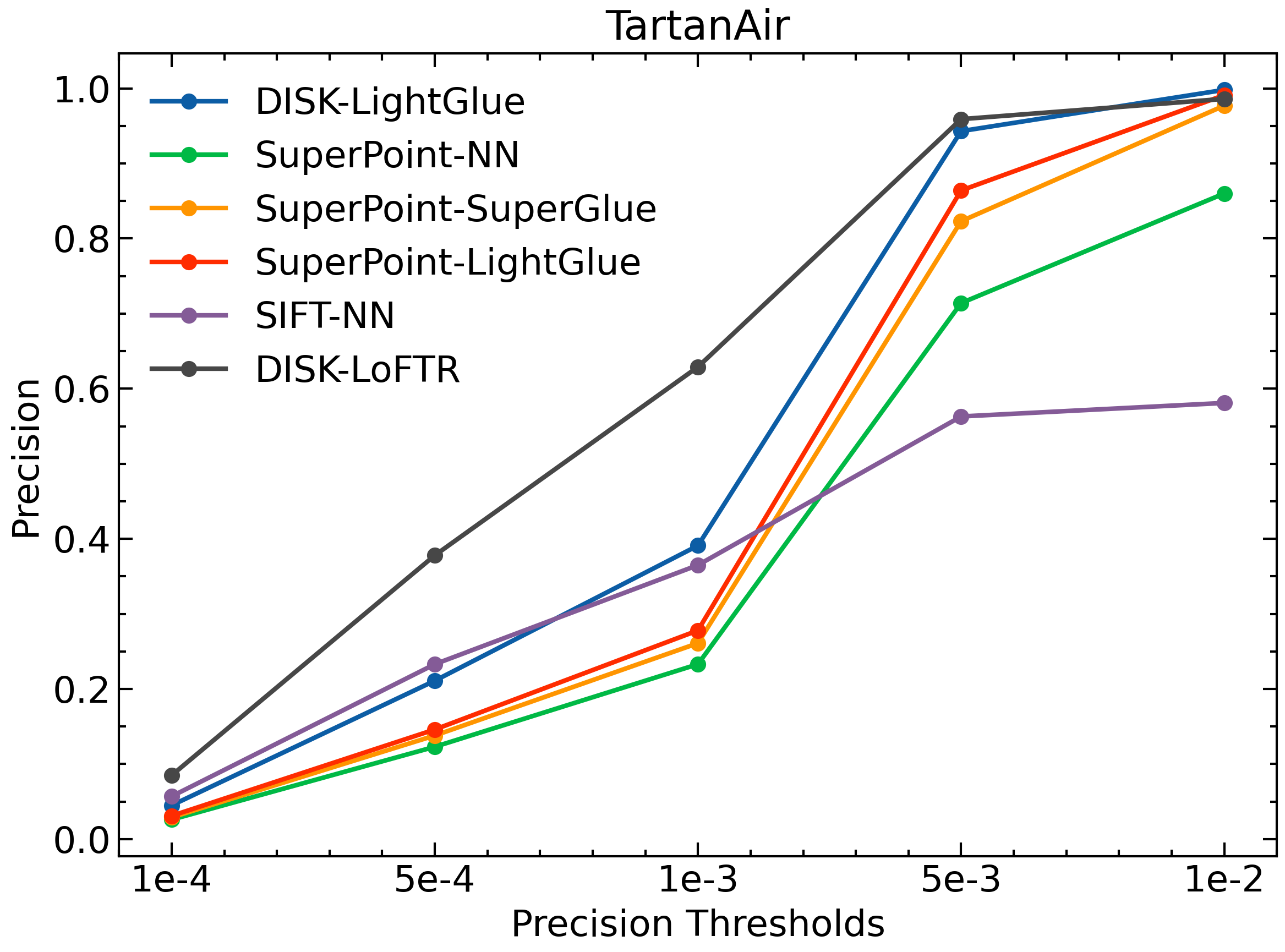}
\caption{\textbf{TartanAir Pre-trained Model Precision:} Deep learning approaches outperformed traditional methods.}
\label{fig:precisionPretrainedTartan}
\end{figure}

\begin{figure}[h!]
\centering
    \includegraphics[width=0.8\linewidth]{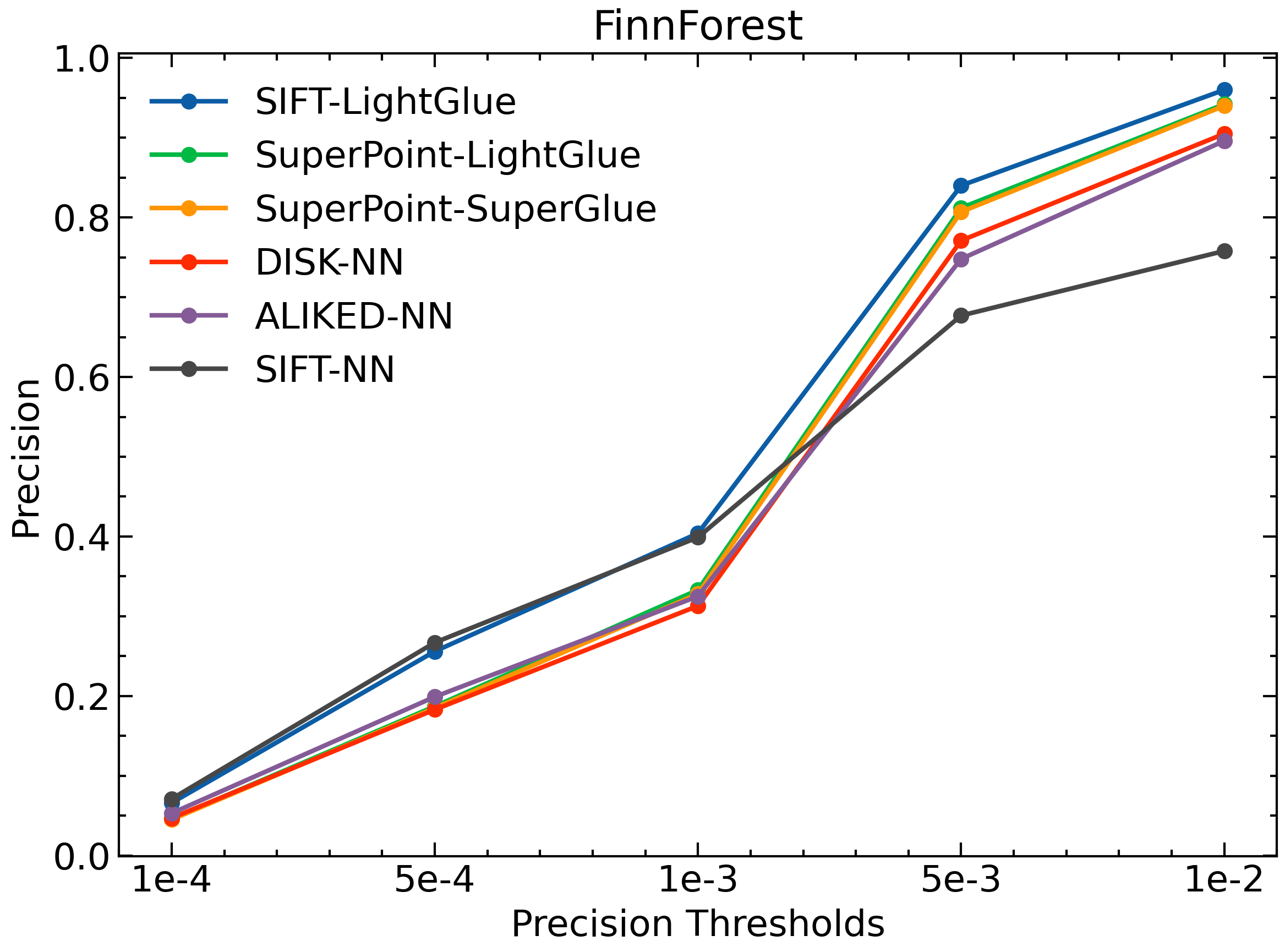}
\caption{\textbf{FinnForest Pre-trained Models' Precision:} SuperGlue and LightGlue showed comparable performance to other learned methods with a decreased computational overhead.}
\vspace{-1.5em}
\label{fig:precisionPretrainedFinn}
\end{figure}

Pre-trained models were assessed on both synthetic, TartanAir \cite{wang2020tartanair}, and real-world, FinnForest \cite{ali2020finnforest}, datasets to compare the precision of deep learning and traditional methods in forest environments (\Cref{fig:precisionPretrainedTartan}, \ref{fig:precisionPretrainedFinn}). Traditional methods like SIFT-NN \cite{lowe2004SIFT, cover1967nearest} exhibited markedly worse precision relative to deep learning approaches, which demonstrated superior generalisation to forest environments despite differences in training domains (\Cref{fig:precisionPretrainedTartan}). 

DISK-LoFTR \cite{tyszkiewicz2020disk,sun2021loftr} achieved the best performance but has a high computational complexity as a dense method that matches every keypoint. In contrast, SuperPoint-LightGlue and SuperPoint-SuperGlue offered comparable precision to both LoFTR and complex extractors like DISK, but with an 80\% decrease in compute time (\Cref{tab:AUCPretrainedTartan}). Similar trends were observed on the real-world FinnForest dataset, where deep learning methods maintained strong performance, with minor precision differences, indicating their potential robustness and effectiveness across different environments (\Cref{fig:precisionPretrainedFinn}).

Locally Optimised RANSAC (LO-RANSAC) \cite{lebeda2012fixing} was used to compute AUC (Area Under Curve) pose estimation scores for relative pose estimation, following the procedure of Lindenberger et al. \cite{lindenberger2306lightglue}. By applying iterative local optimisation steps to refine the inlier set after the initial model fit, LO-RANSAC reduces sensitivity to noise and outliers, improving the robustness and reliability of pose estimation. SuperPoint-LoFTR and SIFT-LoFTR achieved the highest AUC pose estimation scores of 0.916, indicating robust pose estimation, albeit with increased computational demands (\Cref{tab:AUCPretrainedTartan}). In contrast, SuperPoint with LightGlue or SuperGlue processed faster, 88 ms and 142 ms, with comparable pose errors, making them potentially more suitable for real-time, resource-constrained applications. LightGlue provided the best accuracy-efficient trade-off for both precision and pose estimation across both datasets. These findings suggest that SuperGlue and LightGlue are efficient alternatives to dense methods like LoFTR, with the potential for further enhancement through retraining on domain-specific data to increase its accuracy and robustness.

\begin{table}[h!]
\centering
\caption{\textbf{TartanAir Pre-Trained Models' Relative Pose Error:} Dense methods like LoFTR showed the best performance; however, it had a poor precision-efficiency trade off, taking 400\% longer than the equivalent SuperGlue model.}
\label{tab:AUCPretrainedTartan}
\setlength\tabcolsep{1.5pt} 
\newboolean{showRelativeError}
\newboolean{showTiming}  
\setboolean{showRelativeError}{true}     
\setboolean{showTiming}{true}              
\begin{tabular}{cccccc}
\toprule
\multirow{2}{*}{\textbf{Extractor}} & \multirow{2}{*}{\textbf{Matcher}} & \multirow{2}{*}{\textbf{Matches}} &  \multirow{2}{*}{\textbf{Keypoints}} & 
\multicolumn{1}{c}{\textbf{LO-RANSAC AUC}} & \multirow{2}{*}{\ifthenelse{\boolean{showTiming}}{\textbf{Time(ms)}}{}} \\
\cmidrule(lr){5-5}
& & & & \multicolumn{1}{c}{\textbf{5°/10°/20°}} & \\
\midrule 
SuperPoint & LightGlue & 1106&2048  & \ifthenelse{\boolean{showRelativeError}}{0.640/0.745/0.812}{} & \ifthenelse{\boolean{showTiming}}{\bftab88}{} \\
SuperPoint & SuperGlue & 1195&2048 & \ifthenelse{\boolean{showRelativeError}}{0.637/0.744/0.810}{} & \ifthenelse{\boolean{showTiming}}{\bftab142}{} \\
SuperPoint & LoFTR & 3280&3280  & \ifthenelse{\boolean{showRelativeError}}{0.846/0.890/0.916}{} & \ifthenelse{\boolean{showTiming}}{730}{} \\
ALIKED & NN & 1473&2048 & \ifthenelse{\boolean{showRelativeError}}{0.820/0.865/0.894}{} & \ifthenelse{\boolean{showTiming}}{411}{} \\
SIFT & LoFTR & 3280&3280 & \ifthenelse{\boolean{showRelativeError}}{0.846/0.890/0.916}{} & \ifthenelse{\boolean{showTiming}}{800}{} \\
DISK & NN & 1405&2048  & \ifthenelse{\boolean{showRelativeError}}{0.736/0.806/0.851}{} & \ifthenelse{\boolean{showTiming}}{471}{} \\
\bottomrule
\vspace{-3em}
\end{tabular}
\end{table}

\subsection{Forest Glue}
\subsubsection{Refined Models}
The refined models showed improved precision and RANSAC inlier percentages compared to baseline models (\Cref{tab:retrainModelComparisonPrecisionFinn}). The refined LightGlue and SuperGlue models exhibit increased precision, particularly at stricter thresholds below $1e^{-3}$. Both retrained models' RANSAC inlier percentages outperformed the existing models by approximately 10\%, highlighting their greater robustness due to more consistent and accurate correspondences. 

LightGlue’s marginal performance difference compared to SuperGlue indicates that, despite being a lighter version, it achieves comparable performance \cite{lindenberger2306lightglue, sarlin2020superglue}. This may be attributed to its adaptive mechanisms, such as dynamic pruning and early termination, which discard low-confidence matches that contribute minimally to pose estimation accuracy, whilst efficiently retaining key correspondences critical for robust results. Unlike SuperGlue, which applies a fixed attention mechanism to all keypoints, LightGlue dynamically adjusts the number of attention layers based on visual overlap between image pairs. This allows it to reduce unnecessary computations in frames with fewer distinctive features or lower overlap, thereby improving efficiency without compromising accuracy. The model trained from scratch, starting from random weights, showed the lowest performance, likely due to the inherent challenges of training from random initialisation. 

\begin{table}[h!]
\footnotesize
\centering
\caption{\textbf{FinnForest Refined Models' Precision and RANSAC inliers:} Refined models demonstrated higher precision at stricter thresholds and a 10\% increase in RANSAC inliers compared to baseline models.}
\label{tab:retrainModelComparisonPrecisionFinn}
\begin{tabular}{cccc}
\toprule
\multirow{2}{*}{\textbf{Matcher}} & \multirow{2}{*}{\textbf{Training}} & \multicolumn{1}{c}{\textbf{Precision}} & \multirow{2}{*}{\textbf{Inliers \%}} \\ 
\cmidrule(lr){3-3}
& & \multicolumn{1}{c}{\textbf{$1e^{-4}$/$5e^{-4}$/$1e^{-3}$/$5e^{-3}$/$1e^{-2}$}} & \\
\midrule
LightGlue & Scratch & 0.030/0.107/0.190/0.504/0.591 & 21.7 \\ 
LightGlue & Refined & \bftab 0.053/0.214/0.369\normalfont/0.817/0.940 & \textbf{\bftab67.5} \\ 
LightGlue & Baseline & 0.046/0.187/0.333/0.812/0.942 & 57.6 \\ 
SuperGlue & Refined & \bftab 0.052/0.211/0.364\normalfont/0.805/0.932 & \textbf{\bftab66.1} \\ 
SuperGlue & Baseline & 0.045/0.185/0.328/0.807/0.940 & 56.7 \\ 
\bottomrule
\end{tabular}
\end{table}

The refined models achieved similar relative pose errors on the FinnForest dataset using only 512 keypoints (\Cref{tab:AUCRefinedFinnTable}). This suggests these models are well-suited for resource-constrained applications, maintaining performance with fewer keypoints and reduced computational demand. Moreover, the findings indicate that training on synthetic data can enhance performance on real-world forest datasets. Similar trends were observed when evaluating the models on the TartanAir dataset; however, the relative pose errors at the 5° threshold were approximately double those in FinnForest. This suggests that while training with synthetic data improved performance, it may not fully capture the complexities inherent in real-world conditions.

\begin{table}[h!]
\caption{\textbf{FinnForest Refined Models' Pose Estimation Errors:} The refined models showed equivalent relative pose errors despite using 25\% the keypoints of the baseline models.}
\centering
\setlength\tabcolsep{1.5pt} 
\label{tab:AUCRefinedFinnTable}

\newboolean{showRelativeError}
\newboolean{showTranslationalError}
\newboolean{showRotationalError}
\newboolean{showTiming}  

\setboolean{showRelativeError}{true}     
\setboolean{showTranslationalError}{false}  
\setboolean{showRotationalError}{false}    
\setboolean{showTiming}{false}              

\begin{tabular}{cccccc}
\toprule
\multirow{2}{*}{\textbf{Matcher}} &
\multirow{2}{*}{\textbf{Training}} & \multirow{2}{*}{\textbf{KeyPoints}} & \multirow{2}{*}{\textbf{Matches}} & 
\multicolumn{1}{c}{\textbf{LO-RANSAC AUC}} & \multirow{2}{*}{\ifthenelse{\boolean{showTiming}}{\textbf{Time(ms)}}{}} \\
\cmidrule(lr){5-5}
& & & &  \multicolumn{1}{c}{\textbf{5°/10°/20°}} & \\
\midrule
LightGlue & Scratch & 2048 & 942 & \ifthenelse{\boolean{showRelativeError}}{0.004/0.012/0.020}{} \\
LightGlue & Refined & \bftab512 & 282 & \ifthenelse{\boolean{showRelativeError}}{\bftab0.326/0.607/0.788}{} \\
LightGlue & Baseline & 2048 & 1227 & \ifthenelse{\boolean{showRelativeError}}{0.326/0.608/0.789}{} \\
SuperGlue & Refined & \bftab512 & 287 & \ifthenelse{\boolean{showRelativeError}}{\bftab0.325/0.607/0.788}{} \\
SuperGlue & Baseline & 2048 & 1249 & \ifthenelse{\boolean{showRelativeError}}{0.326/0.608/0.789}{} \\
\bottomrule
\end{tabular}
\end{table}

\subsubsection{Multi-Modal Models}
\begin{figure}[h!]
    \centering
    \includegraphics[width=0.9\linewidth]{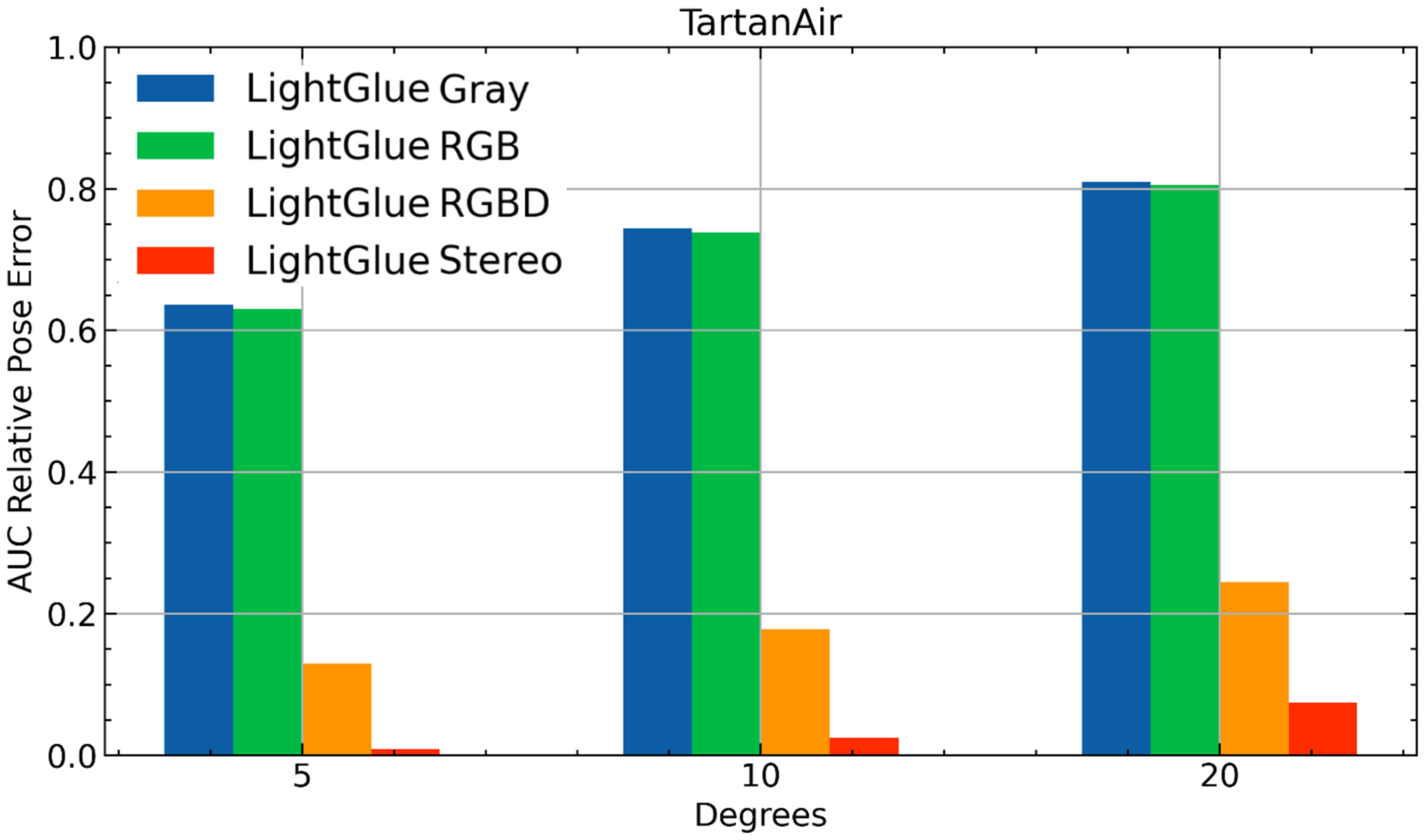}
\caption{\textbf{TartanAir Multi-Modal LightGlue Relative Pose Error:} The relative pose error of RGB and grayscale models showed similar performance. RGB-D and stereo models showed a significant decrease across all thresholds -- Gray (Blue), RGB (Green), RGB-D (Yellow), and Stereo (Red).}
\label{fig:AUCMultiModalTartanlightglue}
\vspace{-1.5em}
\end{figure}

We investigated the effects of multi-modal inputs on performance by extending the input channels of the SuperPoint feature extractor to accommodate RGB, RGB-D, and stereo data in the first convolutional layer. The results demonstrate that RGB and grayscale inputs yield comparable precision and relative pose error performance, with consistent results across SuperGlue and LightGlue models on both datasets (\Cref{fig:AUCMultiModalTartanlightglue}). The maximum relative pose scores for RGB and grayscale models were identical, 0.807 and 0.808, on TartanAir. In contrast, models using RGB-D and stereo inputs demonstrated reduced performance, with maximum relative pose scores reaching 0.244 for RGB-D and 0.074 for stereo on TartanAir, and lower precision across all thresholds. These patterns were consistent across both datasets, indicating that while RGB and grayscale inputs are effective, the current SuperPoint weights are not well-suited for multi-modal inputs, resulting in poor performance.

\subsubsection{Epipolar Models}
We retrained LightGlue with a hinge loss term that penalises predicted matches based on the total epipolar error weighted by their confidence scores.
\begin{table}[h!]
\centering
\caption{\textbf{Epipolar Pose Estimation:} Relative pose error between epipolar and baseline models shows no marked difference on both datasets. However, the epipolar models show an increase in over 20\% match count.}
\label{tab:epipolarAUC}
\setlength\tabcolsep{1.5pt} 
\begin{tabular}{ccccccc}
\toprule
\multirow{2}{*}{\textbf{Matcher}} &  \multirow{2}{*}{\textbf{Dataset}} &
\multirow{2}{*}{\textbf{Mode}} & \multirow{2}{*}{\textbf{KeyPoints}} & \multirow{2}{*}{\textbf{Matches}} & 
\multicolumn{1}{c}{\textbf{LO-RANSAC AUC}} \\
\cmidrule(lr){6-6}
& & & & & \multicolumn{1}{c}{\textbf{5°/10°/20°}} \\
\midrule
 LightGlue & TartanAir & Epipolar & 2048 & \bftab1349 &  0.633/0.739/0.806 \\
 LightGlue & TartanAir & Baseline & 2048 & 1106 & 0.640/0.745/0.812   \\
 LightGlue & FinnForest & Epipolar & 2048 & \bftab1430 & 0.326/0.608/0.789 \\
 LightGlue & FinnForest & Baseline & 2048 & 1227 & 0.326/0.608/0.789 \\ 
\bottomrule
\end{tabular}
\end{table}

The results indicate that this approach generally underperformed compared to the baseline models, with marginally lower precision for both datasets. On the TartanAir dataset, the epipolar modal achieves a precision of 0.239 at the $1e^{-3}$ threshold, 0.04 less than the baseline LightGlue model. The relative pose errors between the epipolar and baseline models remained comparable, suggesting that the epipolar loss did not introduce substantial additional errors in pose estimation (\Cref{tab:epipolarAUC}). Notably, the epipolar models generated over 20\% more matches than the baseline models. This suggests that whilst the matching quantity is increased, it may not enhance the overall model robustness.

\vspace{-0.5em}
\subsection{Pose Estimation Model}
\begin{figure}[h!]
    \centering
    \includegraphics[width=\linewidth]{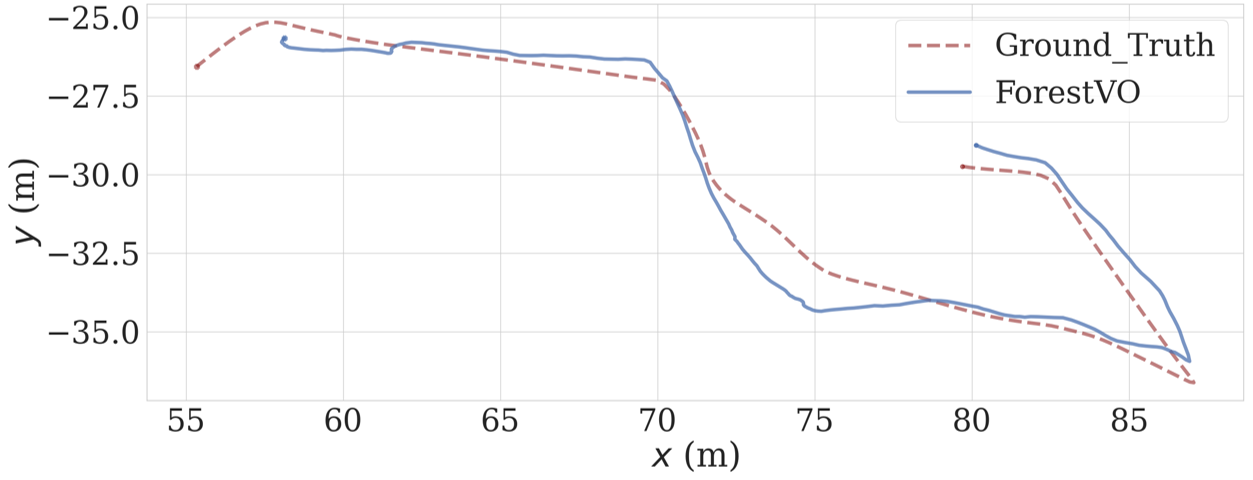}
\caption{\textbf{Estimated and Ground Truth Trajectories:} A visualisation of the \textit{seasonsforest\_sample\_P002} sequence in the TartanAir dataset, comparing the trajectory estimated by our system with the corresponding ground truth. The sequence contains 301 poses and is approximately 50 m in length.}
\label{fig:ForestVO_Traj}
\vspace{-1em}
\end{figure}

\begin{figure}[t!]
    \centering
    \includegraphics[width=0.48\linewidth]{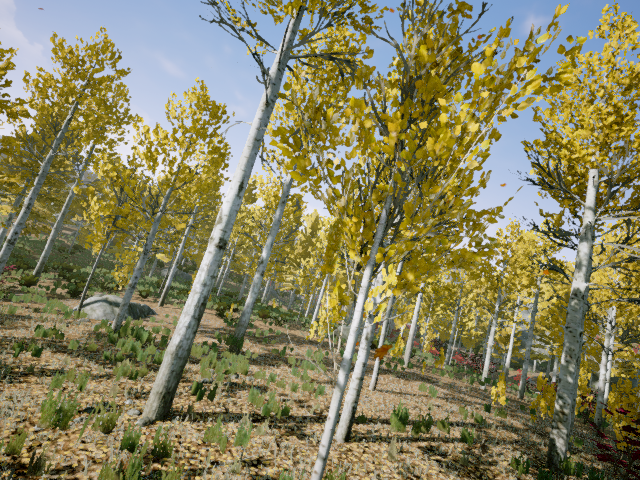}
    \begin{picture}(0,0)
        \put(-15,-10){\small \textbf{(a)}}
    \end{picture}
    \includegraphics[width=0.48\linewidth]{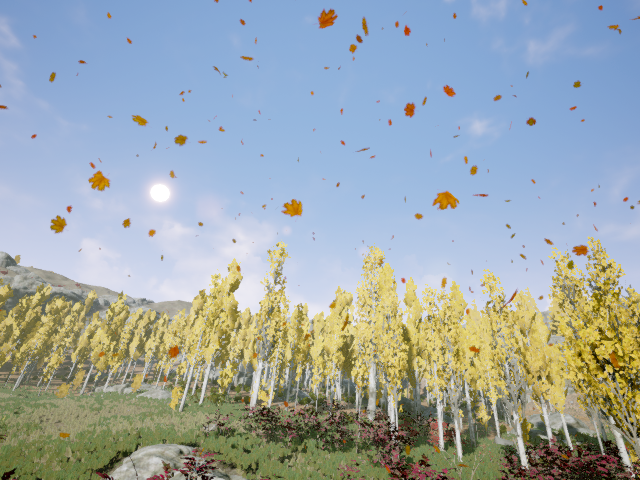}
    \begin{picture}(0,0)
        \put(-15,-10){\small \textbf{(b)}}
    \end{picture}
    \vspace{0em} 
    \caption{\textbf{TartanAir Forest Scenes:} \textbf{(a)} Static \textbf{(b)} Dynamic.}
    \label{fig:TartanAirScenes}
    \vspace{-1em}
\end{figure}
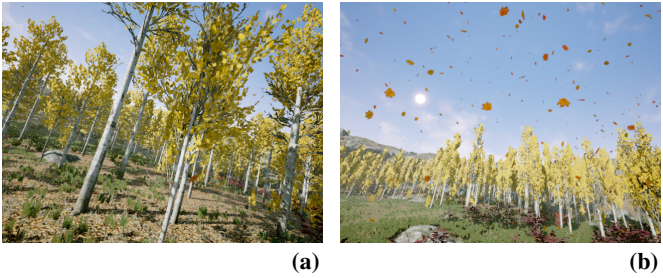

We evaluate our pose estimation model using the Relative Pose Error (RPE) and Absolute Trajectory Error (ATE) metrics on forest sequences from the TartanAir dataset. Fig. \ref{fig:ForestVO_Traj} illustrates one of the trajectory sequences with ground truth and our estimation result. In the dataset, the \textit{easy} scenes contain mainly static objects with limited movement between frames (\Cref{fig:TartanAirScenes}a). The \textit{hard} scenes contain dynamic objects such as falling leaves which create more difficult scenarios for accurate VO (Figure \ref{fig:TartanAirScenes}b). We have multiple options for the feature matcher (SuperGlue \& LightGlue with trained various sensing modalities) and used the grayscale SuperPoint and refined LightGlue models for clarity and brevity in comparisons.

\begin{table}[h]
\centering
\caption{\textbf{TartanAir Relative Pose Estimation Results:} Relative Pose Error (RPE) results of our ForestVO framework on forest scenes from the TartanAir dataset. We use the \textit{rpe\_score} and \textit{kitti\_score} metrics provided by the TartanAir evaluator tool and evaluate using the left images from each sequence. The rpe\_score compares consecutive poses, where \(\Delta t\) is the translational error given in \(m\) and \(\Delta r\) is the rotational error given in \textdegree. The \textit{kitti\_score} \(\Delta t\) represents the average translational error as a \% across the entire trajectory whilst \(\Delta r\) represents the rotational error expressed as \textdegree\(/m\) travelled.}
\label{tab:ForestVOrpe}
\setlength\tabcolsep{1.5pt} 

\begin{tabular}{lcc}
\toprule
\multirow{2}{*}{\textbf{Sequence}} & \multicolumn{1}{c}{\textbf{rpe\_score}} &
\multicolumn{1}{c}{\textbf{kitti\_score}} \\
\cmidrule(lr){2-2}
\cmidrule(lr){3-3}
& \multicolumn{1}{c}{\textbf{\(\Delta t (m)\) \(\Delta r\)(\textdegree)}} & \multicolumn{1}{c}{\textbf{\(\Delta t (\%)\) \(\Delta r\)(\textdegree\(/m\))}} \\
\midrule
Gascola\_Easy\_P008 & 1.895 / 32.591 & 3.448 / 0.365 \\
Gascola\_Hard\_P009 & 1.244 / 19.437 & 1.742 / 0.275 \\
Seasons\_Forest\_Easy\_P011 & 0.952 / 6.275 & 2.724 / 0.250 \\
Seasons\_Forest\_Hard\_P006 & 0.781 / 4.460 & 2.490 / 0.349 \\
Seasons\_Forest\_Winter\_Easy\_P009 & 0.816 / 8.032 & 2.114 / 0.282 \\
Seasons\_Forest\_Winter\_Hard\_P018 & 0.853 / 17.57 & 1.482 / 0.251 \\
\midrule
Average & 1.090 / 14.73 & 2.333 / 0.295 \\
\bottomrule
\end{tabular}
\vspace{-1em}
\end{table}

\Cref{tab:ForestVOrpe} shows the \textit{rpe} and \textit{kitti} scores for the forest testing sequences. The relative translations are quite similar when comparing the \textit{easy} and \textit{hard} sequences which suggests that the model is robust to the dynamic objects present in the \textit{hard} scenes. The Rotational error fluctuates more between consecutive frames, likely because TartanAir’s relative rotations are smaller than translations, resulting in the model better predicting translations.
The scaling factor \(\beta\) = 100 which was used to balance the translation and rotation losses when training the model may not be optimal and should be fine-tuned further.

\begin{table}[h!]
\centering
\caption{\textbf{Direct vs Indirect Methods:} Absolute Trajectory Error (ATE) and RPE comparisons on forest scenes from the TartanAir dataset. We compare against the DSO-Monocular direct method using the \textit{easy} and \textit{hard} sequences from the autumn and winter forest scenes in the TartanAir dataset. We reduce the length of each sequence to 200 frames to match the format used in \cite{wang2020tartanair}.}
\label{tab:ta_easy_vs_hard}
\setlength\tabcolsep{1.5pt} 

\begin{tabular}{lcc}
\toprule
\multirow{2}{*}{\textbf{Environment}} & \multicolumn{1}{c}{\textbf{ForestVO}} & \multicolumn{1}{c}{\textbf{DSO}} \\
\cmidrule(lr){2-2}
\cmidrule(lr){3-3}
& \multicolumn{1}{c}{\(ATE(m)\) \(\Delta t (m)\) \(\Delta r\)(\textdegree)}
& \multicolumn{1}{c}{\(ATE(m)\)  \(\Delta t (m)\) \(\Delta r\)(\textdegree)} \\
\midrule
Autumn + Static & 2.010/0.542/3.335 & \textbf{1.660}/\textbf{0.016}/\textbf{1.092} \\
Autumn + Dynamic & \textbf{2.553}/\textbf{0.535}/\textbf{2.739} & */*/* \\
Winter + Static & \textbf{1.517}/0.644/4.040 & 2.305/\textbf{0.196}/\textbf{1.745} \\
Winter + Dynamic & \textbf{1.718}/\textbf{0.512}/\textbf{3.819} & 3.707/0.870/4.566 \\
\bottomrule
\end{tabular}
\end{table}

Table \ref{tab:ta_easy_vs_hard} compares the performance of ForestVO with DSO-Monocular on the \textit{easy} and \textit{hard} scenes. DSO is a direct method which uses optical flow for dense pixel matches instead of sparse feature matching. DSO achieves lower RPE scores in the static environments but performs much worse when motion artifacts such as moving leaves are introduced. In dynamic scenes, the ATE score for ForestVO increases by approximately 20\% whereas DSO increases by approximately 60\%. This suggests that direct methods are unsuitable for challenging forest scenes since the optical flow method confuses the movement of leaves with the camera movement, resulting in less accurate relative pose estimations. On the contrary, our method relies on ForestGlue, a robust feature matcher which can better handle dynamic scenes and therefore the performance drop from the \textit{easy} scenes to the \textit{hard} scenes is marginal in our case.

\begin{table}[h!]
\centering
\caption{\textbf{TartanAir ATE Results:} ATE comparisons on the TartanAir monocular sequences against existing VO/SLAM methods. The ATE scores of the other methods are averaged across all sequences. However, since our method is only trained on forest scenes, we compute the average ATE score for our method using only the forest scenes from the TartanAir dataset.}
\label{tab:ForestVOATE}
\setlength\tabcolsep{1.5pt}

\begin{tabular}{ccccc}
\toprule
\textbf{Method} & ORB-SLAM & DeepV2D & TartanVO & Ours \\
\midrule
\(ATE(m)\) & 5.06 & 5.03 & \textbf{1.92} & 1.95 \\
\bottomrule
\end{tabular}
\end{table}

Table \ref{tab:ForestVOATE} compares the ATE of different VO/SLAM methods on the TartanAir dataset. Despite no loop closure detection or bundle adjustment in our system, ForestVO outperforms the ORB-SLAM and Deepv2D methods. It is only marginally worse than the TartanVO method, however, we only trained the pose estimation model on the forest scenes, which make up approximately 10\% of TartanAir dataset, amounting to only 75000 image pairs, whereas TartanVO was trained on the entire dataset.

\section{Conclusion}
We have demonstrated that deep learning-based approaches outperformed the traditional methods in complex forest environments, where the challenges of unstructured settings necessitate robust feature matching and pose estimation. LightGlue was shown to effectively compromise computational efficiency and accuracy, making it promising for real-time applications in resource-constrained environments. Incorporating domain-specific retraining further improved model performance, indicating that adaptation to the specific characteristics of forest environments can lead to notable enhancements. ForestGlue achieves comparable pose estimation accuracy to baseline LightGlue and SuperGlue models while using only 512 keypoints -- just 25\% of the 2048 keypoints required by baseline models -- to achieve a LO-RANSAC AUC score of 0.745 at a 10\textdegree\ threshold, highlighting its computational efficiency. Moreover, the potential for generalising from synthetic data to real-world applications was explored, suggesting that models can effectively transition between these domains with appropriate retraining.

Having a robust feature matcher for forest environments facilitated the training of a novel pose estimation model which directly uses the 2D matched keypoints coordinates to regress the relative camera pose between a pair of frames. On challenging TartanAir forest sequences, our ForestVO framework achieved an average RPE of 1.09 m and kitti\_score of 2.33\%, outperforming direct-based VO methods such as DSO in dynamic scenes by 40\%, with an ATE of 1.718 m. ForestVO maintains competitive performance with TartanVO despite training on only 10\% of the TartanAir dataset. We have demonstrated the robustness of this framework on challenging forest scenes, highlighting the potential for such lightweight models being utilised for VO applications in forest environments.
\vspace{-0.2em}

\section{Future Work}
While synthetic datasets such as TartanAir have been invaluable for developing domain-adapted models for forest environments, their applicability to real-world scenarios is hindered by the domain gap between virtual and natural scenes. Real-world datasets would close this domain gap, however, refining model weights requires both depth and pose annotations, which no known forest dataset currently provides. To address this, we propose the collection of a dedicated real-world dataset using a gimbal-mounted RGB-D camera on a drone, alongside IMU and RTK-GPS sensors for accurate ground-truth pose estimation. To ensure generalisation, this dataset should capture woodlands of varying types across seasons, weather conditions, and lighting variations. It must also account for real-world challenges that are less common or poorly represented in synthetic datasets, such as lens flares, dynamic tree movement, and changes in foliage density. Incorporating real-world data into the training of LightGlue and SuperGlue facilitates more effective domain adaptation, enabling the models to effectively capture the complexities of natural environments and achieve greater robustness in real-world deployments.

Beyond collecting datasets, the system must undergo real-world testing through outdoor drone flights using ForestVO for pose estimation.
These field tests will comprehensively evaluate real-time performance, addressing computational efficiency, latency, and resilience to environmental changes. Combining real-world data for training with practical drone deployments will ensure that the models can accurately and reliably perform visual odometry in unstructured forest environments by enhancing their applicability to real-world autonomous navigation tasks.

The future work would need to evaluate the real time performance on resource constraint platforms with comprehensive benchmarking of the proposed framework, in particular, a diverse set of real-world hardware systems commonly used in mobile robotics. This will include evaluating performance on various processors, embedded systems, and robotic platforms to assess scalability, efficiency, and practical usability.

\bibliographystyle{IEEEtran}
\bibliography{bibliography}

\end{document}